\documentclass[letterpaper, 10 pt, conference]{00_config/ieeeconf}  % Comment this line out if you need a4paper
\IEEEoverridecommandlockouts                              % Needed to use the \thanks command
\usepackage[USenglish]{babel}
\usepackage{csquotes}

\usepackage{amsmath,amssymb,amsfonts}
\usepackage{algorithm,algorithmic}
\usepackage{graphicx}
\usepackage{textcomp}

\usepackage[
        style=ieee,             % numerischer style
        backend=bibtex,          % Bibtex-Format
        isbn=false,             % Unterdrückung ISBN-Angabe
        doi=false,              % Unterdrückung DOI-Angabe
        giveninits=true,        % Nur Initialien der Vorname
        url=false,              % no url
        date=year,              % nur Jahr als Datumsanzeige
        maxbibnames=99,         % maximale Anzahl an Autoren im Literaturverzeichnis
        minnames=1,             % minimale Anzahl an Autoren bei der Zitierung
        maxnames=5,             % maximale Anzahl an Autoren bei der Zitierung
        sorting=none,           % nach Auftreten im Text
        alldates=year
]{biblatex}

\usepackage[nonumberlist,nomain,nogroupskip,section=section,automake]{glossaries}

\usepackage{hyphenat}

\usepackage{mathtools}  % for bmatrix* alignment
\usepackage{bm}         % bold and italic for symbols
\usepackage{siunitx}	% Measurement Units
\usepackage{xkeyval}        % Needed for todonotes
\usepackage{todonotes}		% Dependencies: ifthen, xkeyval, xcolor, tikz, calc, graphicx

\usepackage{hyphenat}           % to hyphenate words that are already hyphenated
\usepackage{enumerate}
\usepackage{paralist}

\usepackage{tikz}
\usepackage{pgfplots}
\usetikzlibrary{external}
\pgfplotsset{compat=newest}
\usepgfplotslibrary{fillbetween}
\usetikzlibrary{math} %needed tikz library
\usetikzlibrary{shapes.geometric, arrows, positioning}

\usepackage{subcaption}
\usepackage{hyperref}
\hypersetup{hidelinks}

\usepackage{eso-pic}

\newglossary[alg]{abb}{asi}{aso}{Abbreviations}
\makeglossaries

\addto\extrasUSenglish{%
}

\newcommand\submittedtext{%
  \footnotesize This work has been submitted to the IEEE for possible publication. Copyright may be transferred without notice, after which this version may no longer be accessible.}

\newcommand\submittednotice{%
  \AddToShipoutPictureBG*{%
    \begin{tikzpicture}[remember picture,overlay]
      \node[anchor=south,yshift=13pt] at (current page.south) {\parbox{0.65\textwidth}{\centering\submittedtext}};
    \end{tikzpicture}%
  }%
}

\newacronym{ODE}{ODE}{ordinary differential equation}
\newacronym{COG}{COG}{center of gravity}
\newacronym{ROS}{ROS}{Robot Operating System}
\newacronym{STO}{STO}{Safe Torque Off}
\newacronym{SOEM}{SOEM}{Simple Open EtherCAT Master}
\newacronym{CSV}{CSV}{Cyclic Synchronous Velocity}
\newacronym{CSP}{CSP}{Cyclic Synchronous Position}
\newacronym{MPC}{MPC}{Model Predictive Control}

\newacronym{AV}{AV}{autonomous vehicles}
\newacronym{GNSS}{GNSS}{Global Navigation Satellite Systems}
\newacronym{SLAM}{SLAM}{Simultaneous Localization and Mapping}
\newacronym{IMU}{IMU}{Inertial Measurement Unit}
\newacronym{EKF}{EKF}{Extended Kalman Filter}
\newacronym{PDO}{PDO}{Process Data Object}
\newcommand{\vF}{\bm{F}}      % force vector
\newcommand{\sF}{F}           % force scalar
\newcommand{\sM}{M}           % torque scalar
\newcommand{\vG}{\bm{G}}      % tire chassis coupling matrix
\newcommand{\vR}{\bm{R}}      % tire chassis coupling matrix

\newcommand{\vv}{\bm{v}}      % velocity vector
\newcommand{\va}{\bm{a}}      % velocity vector

\newcommand{\overleftsmallarrow}{\mathpalette{\overarrowsmall@\leftarrowfill@}}
\newcommand{\overarrowsmall@}[3]{%
  \vbox{%
    \ialign{%
      ##\crcr
      #1{\smaller@style{#2}}\crcr
      \noalign{\nointerlineskip}%
      $\m@th\hfil#2#3\hfil$\crcr
    }%
  }%
}
\def\smaller@style#1{%
  \ifx#1\displaystyle\scriptstyle\else
    \ifx#1\textstyle\scriptstyle\else
      \scriptscriptstyle
    \fi
  \fi
}
\makeatother

\newcommand{\RNum}[1]{\uppercase\expandafter{\romannumeral #1\relax}}
\newcommand{\zelos}{\textit{ZeloS}}

\newcommand{\abs}[1][]{\lvert #1 \rvert}        % absolute

\newcommand{\transpose}{^\top}             % transpose

\newcommand{\rBrack}[1][]{\left( #1 \right)}    % round brackets
\newcommand{\cBrack}[1][]{\left\{ #1 \right\}}  % curly brackets

\newcommand{\vf}{\bm{f}}                   % vectorized function

\newcommand{\diff}{\mathrm{d}}                              % Differential
\newcommand{\dFull}[2]{\dfrac{\diff #1}{\diff #2}}		% Full derivative
\newcommand{\dFullInline}[2]{{\diff #1}/{\diff #2}}		% Full derivative inline

\newcommand{\indP}{_\mathrm{p}}         % index planner
\newcommand{\indPA}{_\mathrm{p,a}}         % index planner admissible
\newcommand{\indRunI}{_{i}}             % running index i
\newcommand{\indx}{_\mathrm{x}}         % index x
\newcommand{\indy}{_\mathrm{y}}         % index y
\newcommand{\indz}{_\mathrm{z}}         % index z
\newcommand{\indo}{_\mathrm{\so}}         % index z

\newcommand{\indxref}{_\mathrm{\sx,r}}
\newcommand{\indyref}{_\mathrm{\sy,r}}
\newcommand{\indxdes}{_\mathrm{\sx,d}}
\newcommand{\indydes}{_\mathrm{\sy,d}}
\newcommand{\indxoff}{_\mathrm{\sx,o}}
\newcommand{\indyoff}{_\mathrm{\sy,o}}

\newcommand{\indcog}{_\mathrm{cog}}
\newcommand{\indxy}{_\mathrm{xy}}
\newcommand{\indRunIx}{_{i,\mathrm{x}}}
\newcommand{\indRunIy}{_{i,\mathrm{y}}}
\newcommand{\indfl}{_\mathrm{fl}}
\newcommand{\indfr}{_\mathrm{fr}}
\newcommand{\indrl}{_\mathrm{rl}}
\newcommand{\indrr}{_\mathrm{rr}}
\newcommand{\indref}{_\mathrm{r}}
\newcommand{\inddes}{_\mathrm{d}}
\newcommand{\indoff}{_\mathrm{o}}

\newcommand{\indeps}{_\mathrm{\varepsilon}}

\newcommand{\vu}{\bm{u}}                  % input vector
\newcommand{\vx}{\bm{x}}                  % state vector
\newcommand{\vxdot}{\dot{\vx}}            % state vector dot
\newcommand{\sx}{x}                       % state x
\newcommand{\sy}{y}                       % state y
\newcommand{\sv}{v}                       % state v
\newcommand{\serr}{\varepsilon}                  % state vector
\newcommand{\so}{\varphi}                 % state orientation
\newcommand{\sa}{a}                       % scalar acceleration
\newcommand{\sdo}{\dot{\so}}              % scalar yaw rate
\newcommand{\sddo}{\ddot{\so}}              % scalar yaw rate
\newcommand{\stime}{t}                      % time
\newcommand{\reals}[1][]{\mathbb{R}^{ #1 }}     % reals of dimension #1

\newcommand{\admissX}{\mathcal{X}}     % set of admissible states
\newcommand{\admissU}{\mathcal{U}}     % set of admissible inputs

\title{\LARGE \bf
\zelos\ -- A Research Platform for Early-Stage Validation of Research Findings Related to Automated Driving% the Added Value of Ensuring Safety by Design
}

\author{Christopher Bohn$^{1}$, Florian Siebenrock$^{1}$, Janne Bosch$^{1}$, Tobias Hetzner$^{1}$, Samuel Mauch$^{1}$, Philipp Reis$^{2}$,\\ Timo Staudt$^{1}$, Manuel Hess$^{1}$, Ben-Micha Piscol$^{1}$, and Sören Hohmann$^{1}$% <-this % stops a space
\thanks{$^{1}$ Authors are with the Institute of Control System (IRS) at the Karlsruhe Institute of Technology (KIT), 76131 Karlsruhe, Germany.}%
\thanks{$^{2}$ Author is with the Forschungszentrum Informatik (FZI), 76131 Karlsruhe, Germany.}%
\thanks{Corresponding author is C. Bohn, {\tt\small christopher.bohn@kit.edu}.}%
}

\begin{document}

% Frame
\maketitle
\thispagestyle{empty}
\pagestyle{empty}
\begin{abstract}
    This paper presents \zelos, a research platform designed and built for practical validation of automated driving methods in an early stage of research. We overview \zelos' hardware setup and automation architecture and focus on motion planning and control. \zelos\ weighs 69\,kg, measures a length of 117\,cm, and is equipped with all-wheel steering, all-wheel drive, and various onboard sensors for localization. The hardware setup and the automation architecture of \zelos\ are designed and built with a focus on modularity and the goal of being simple yet effective. The modular design allows the modification of individual automation modules without the need for extensive onboarding into the automation architecture. As such, this design supports \zelos\ in being a versatile research platform for validating various automated driving methods. The motion planning component and control of \zelos\ feature optimization-based methods that allow for explicitly considering constraints. We demonstrate the hardware and automation setup by presenting experimental data.
\end{abstract}

\submittednotice

% Content
\section{Introduction}
Practical test platforms are crucial to the research and development of automated driving methods \cite{werling.2017}. Unlike simulation\hyp{}based validations, practical test platforms confront new methods with the unpredictable complexities of a real-world environment. Often, this raises awareness of potentially unrecognized matters. For these reasons, practical tests should be facilitated in an early stage in the development process for goal-oriented refinement and optimization \cite{porter.2004}. 

\begin{figure}[t!]
    \centering
    \includegraphics[width=\columnwidth]{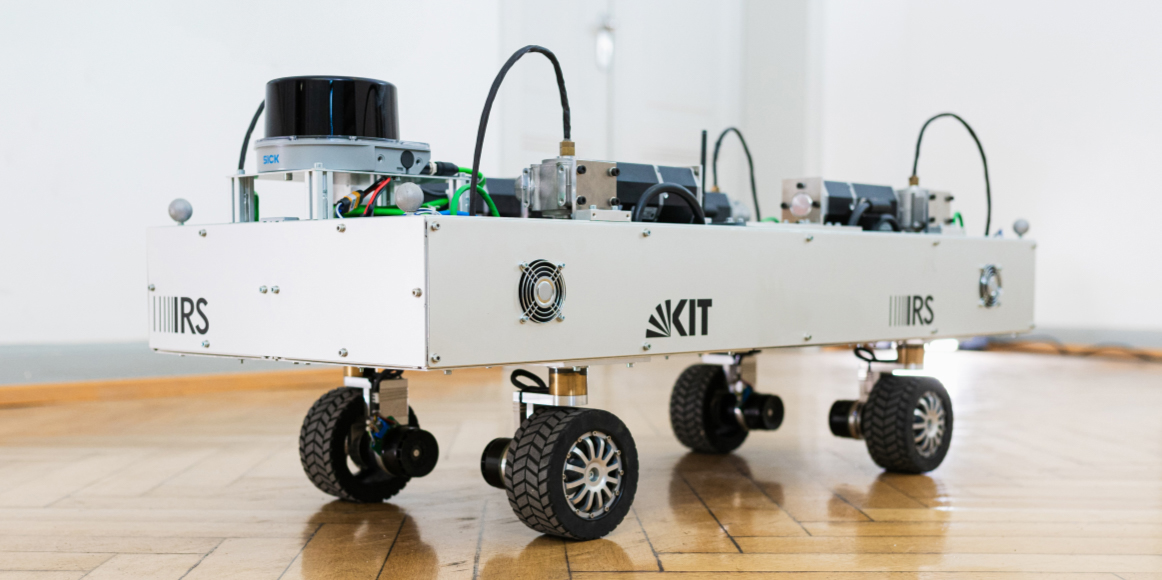}
    \caption{\zelos, a versatile research platform.}
    \label{fig:ZeloS-picture}
\end{figure}

In this paper, we present \zelos, a research platform that allows for early-stage testing and validation of research findings related to automated driving, see \autoref{fig:ZeloS-picture}. With this paper, we aim to encourage other research groups to facilitate the benefits of practical validations in an early stage. To this end, we share our insights from designing and building the research platform and present an overview of the hardware setup and automation architecture.

\zelos\ serves as a versatile research platform, designed and built for validating various automated driving methods, rather than being a purpose-built test vehicle. To ensure flexibility, the hardware and automation of \zelos\ are designed to be modular with a simple yet effective architecture. \zelos\ is equipped with all-wheel steering and all-wheel drive, which allows switching between front-wheel steering, rear-wheel steering, or all-wheel steering. In addition, \zelos\ features powerful actuators and state-of-the-art onboard sensors. %The implemented performance is greater than initially required, allowing for scaling back if needed while maintaining sufficient reserves for demanding validations. 
The modular automation architecture supports modifying and replacing individual automation modules for seamless validation of various automation components. \zelos\ measures a length of $ l = \SI{117}{\centi\meter} $, weighs $ m = \SI{69}{\kilogram} $, and features an electrical power supply at $ V = \SI{48}{\volt} $. This scale allows for easy handling of \zelos, reducing risks, efforts, and costs of practical validations while enabling compatibility with professional sensors and actuators. In addition, a digital twin of \zelos\ enables fast and risk-free software in the loop testing of automation algorithms.

Our initial intent of using \zelos\ is to validate motion planning and control algorithms that formally ensure functional safety \cite{ISO26262}. Formally ensuring safety aims to reduce the effort required to demonstrate safety through large-scale testing. However, embedding safety considerations directly into the design process increases the complexity of the design process and, consequently, necessitates early-stage validation. % of the applied methods.% This area of research is where the name \zelos\ is rooted: inspired by Greek mythology, the name \zelos\ symbolizes the \textit{dedicated and passionate pursuit} of formally ensuring the \textit{safety} of automated systems.
\subsection{Review on Existing Concepts}
Existing vehicle demonstrators can be categorized into three groups: purpose-built test vehicles, vehicles built for automated driving competitions, and vehicles that serve as research platforms. Vehicles of the different categories are designed with distinct objectives and characteristics.

Purpose-built test vehicles are designed to test a specific method, system, or technology in a real-world environment. For this reason, it is often difficult to test methods other than those for which the system was initially designed to test. Purpose-built test vehicles are, e.g., the MARTY demonstrator \cite{stanford} and the EDGAR demonstrator \cite{edgar}. MARTY is designed to test control algorithms that handle a vehicle close to its dynamic driving limits; EDGAR features a design emphasis on redundancy and operational reliability. Moreover, most small-scale demonstrators reviewed in \cite{small-scale} are designed to test a specific automation system in a cost-effective and low-risk environment.

Vehicles built specifically for competitions are designed with a focus on performance while meeting competition requirements. Often, embedded automation algorithms are used that impede the modification of an individual automation system. Examples are the F1TENTH autonomous driving platform \cite{f1tenth}, and Formula Student vehicles, e.g., AMZ Racing \cite{kabzan.2020}.

In contrast, research platforms are versatile environments for various validation tasks. For example, the vehicle in \cite{wei.2013} is built for testing a whole automation architecture and different methods for driving on public roads. Another example of this category is the CoCar NextGen demonstrator \cite{fzi_cocar}, which provides a modular architecture that enables flexible setups for different automated driving-related research.

\zelos\ is designed and built as a research platform. In contrast to \cite{wei.2013,fzi_cocar}, \zelos\ benefits from the smaller scale, rendering it a practical solution in research validations with financial, spatial, and human safety considerations.

\section{Setup of the Research Platform}
In this section, we present the hardware setup, automation architecture, and the safety concept of \zelos.
\subsection{Overview}
\zelos\ measures a length of $ l = \SI{117}{\centi\meter} $ and weighs $ m = \SI{69}{\kilogram} $. The vehicle body is designed to be modular; it is composed of five hardware modules: one center module and four drive modules (see \autoref{fig:zelos_cad}). This design offers flexibility as the arrangement of the wheels can easily be modified. In addition to the modular design of the hardware, the automation is composed of modular software units. As such, it provides a simple yet effective and flexible architecture with minimal potential sources of error. The safety concept is implemented close to steering and driving actuators, so modifying automation modules at higher automation layers does not impede the safe operation of \zelos.

\subsection{Vehicle Body Design}\label{sec:body_design}
\begin{figure}[t!]
    \centering
    \includegraphics[        
    width=\columnwidth,
    height=3.75cm,
    keepaspectratio,]{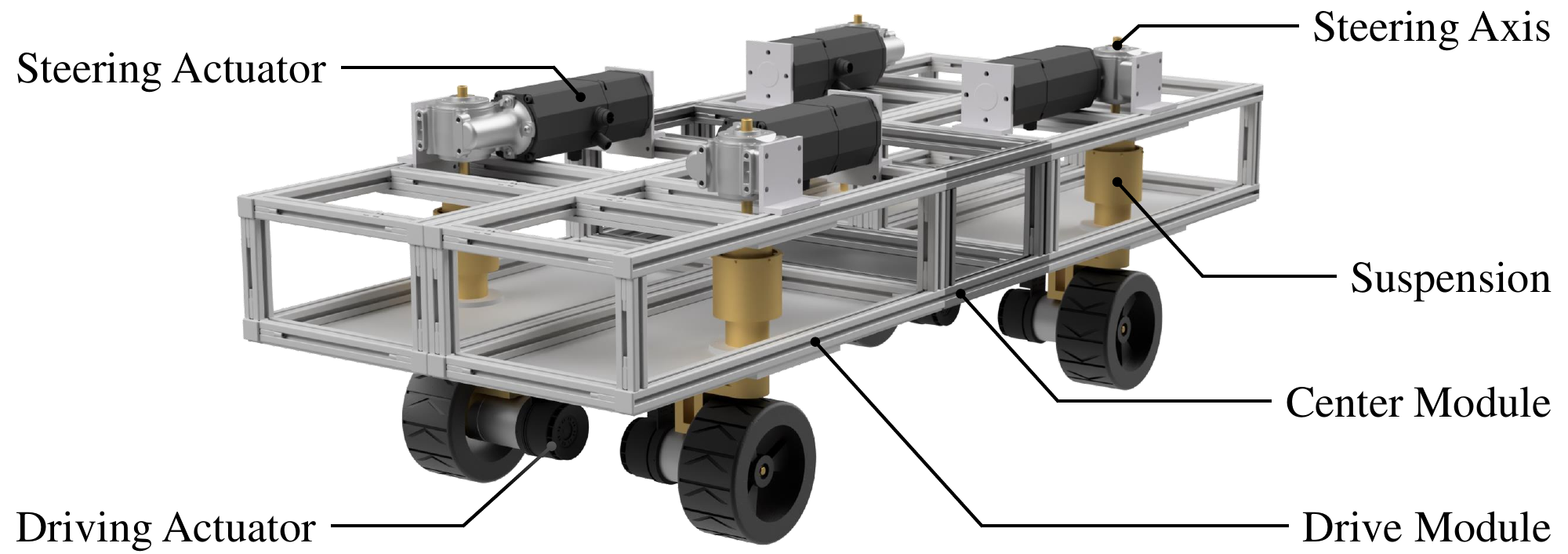}
    \caption{The CAD drawing of the vehicle body reveals the modular design.}
    \label{fig:zelos_cad}
\end{figure}

Each hardware module functions as a self-contained unit, housed within a square aluminum profile frame (see \autoref{fig:zelos_cad}). Every hardware module is equipped with an independent power supply and a computational unit. Additionally, the drive modules include a steering and a driving actuator. Thus, \zelos\ is equipped with all-wheel steering and all-wheel drive. This design offers flexibility, scalability, and maintainability.

The chassis and suspension design is optimized for simple mathematical modeling rather than maximizing driving performance or comfort. This is a key advantage of \zelos\ as its horizontal dynamics on a flat surface can be accurately modeled using a simple mathematical model, which in turn simplifies calculations for motion planning and control.

The steering axes are parallel to the $ z $-axes of the vehicle body and located exactly above the tire contact points. Thus, the tire contact points remain steady w.r.t. \zelos' \gls{COG} for arbitrary tire steering angles. Moreover, each drive module is equipped with a linear suspension so that the tire contact points remain steady for different deflection ranges of the suspension. 
\subsection{Driving and Steering Actuators}\label{sec:actuators}
All drive modules are equipped with a Maxon EC 60-200 driving actuator (controlled by an EPOS4 controller) and a Dunkermotoren BG66x25 dPro steering actuator. Each driving actuator features a maximum driving power of $ P_\mathrm{\omega,max} = \SI{1}{\kilo\watt} $ (overload operation), which is sufficient to break traction on all four tires from a standstill. Each steering actuator features a maximum output power of $ P_\mathrm{\delta,max} = \SI{277}{\watt} $, which allows for reaching a desired tire steering angle almost instantly. This level of performance ensures sufficient reserves to support demanding vehicle validations, particularly for dynamic control and planning scenarios.
\subsection{Hardware Interface}\label{sec:hw_interface}
Each drive module runs a custom-developed drive control unit that connects the controllers of the driving and steering actuators to the automation. The drive control unit is based on the open-source \gls{SOEM} library. It manages the EtherCAT state machine and uses the CiA402 motion profile to control and monitor the steering and driving actuators. The drive control unit operates with an EtherCAT cycle rate of $ \SI{1}{\kilo\hertz} $ and uses the \gls{CSV} mode for the driving actuators and the \gls{CSP} mode for the steering actuators. 

Beyond essential motion control, the drive control unit internally abstracts the CiA402 state machine to simplify communication with the safety unit. Therefore, the safety unit only requires one enable operation interface and one safe stop interface per drive module. In doing so, the hardware interface enforces critical safety functions long before higher-level safety or monitoring components are involved.
\subsection{Computational Units}
\begin{figure}[t!]
    \centering
    \includegraphics[
        width=0.75\columnwidth,
        height=3.75cm,
        keepaspectratio,
      ]{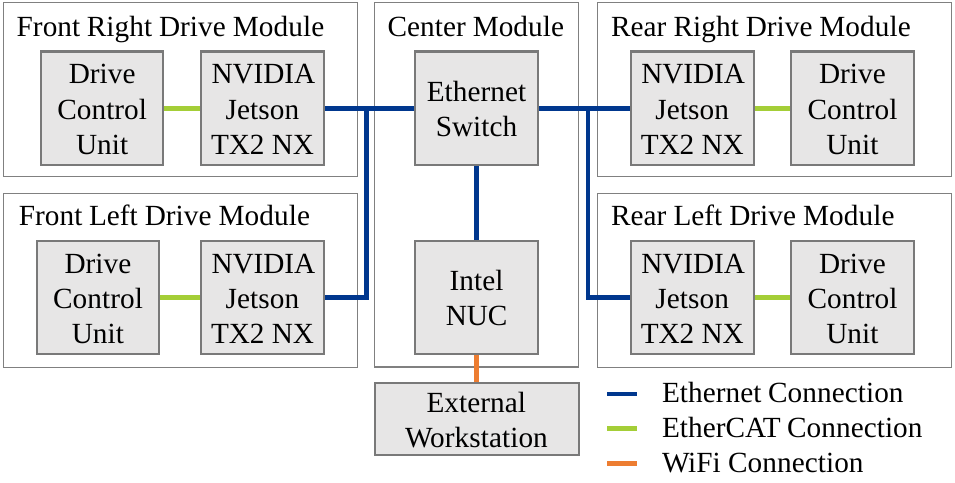}
      \caption{Hardware Architecture of \zelos.}
      \label{fig:cpu_architecture}
\end{figure}

\zelos\ features multiple onboard computational units; the hardware architecture is depicted in \autoref{fig:cpu_architecture}. The main onboard CPU is an Intel NUC, located on the center module. The Intel NUC communicates wirelessly via Wi-Fi with an external workstation that provides an interface for the user to monitor \zelos' systems and to generate validation and test scenarios. In addition, each drive module is equipped with an NVIDIA Jetson TX2 NX that is connected to the Intel NUC.

The Intel NUC provides sufficient computational performance for hosting all automation modules of a centralized automation architecture. Each NVIDIA Jetson TX2 NX serves as an interface to the driving and steering actuators by hosting a drive control unit. Moreover, the NVIDIA Jetson TX2 NX provide sufficient computing performance to host automation modules if a decentralized automation architecture needs to be tested or validated.
\subsection{Automation Architecture}
\begin{figure}[t!]
    \centering
    \includegraphics[
        width=\columnwidth,
        height=6cm,
        keepaspectratio,
      ]{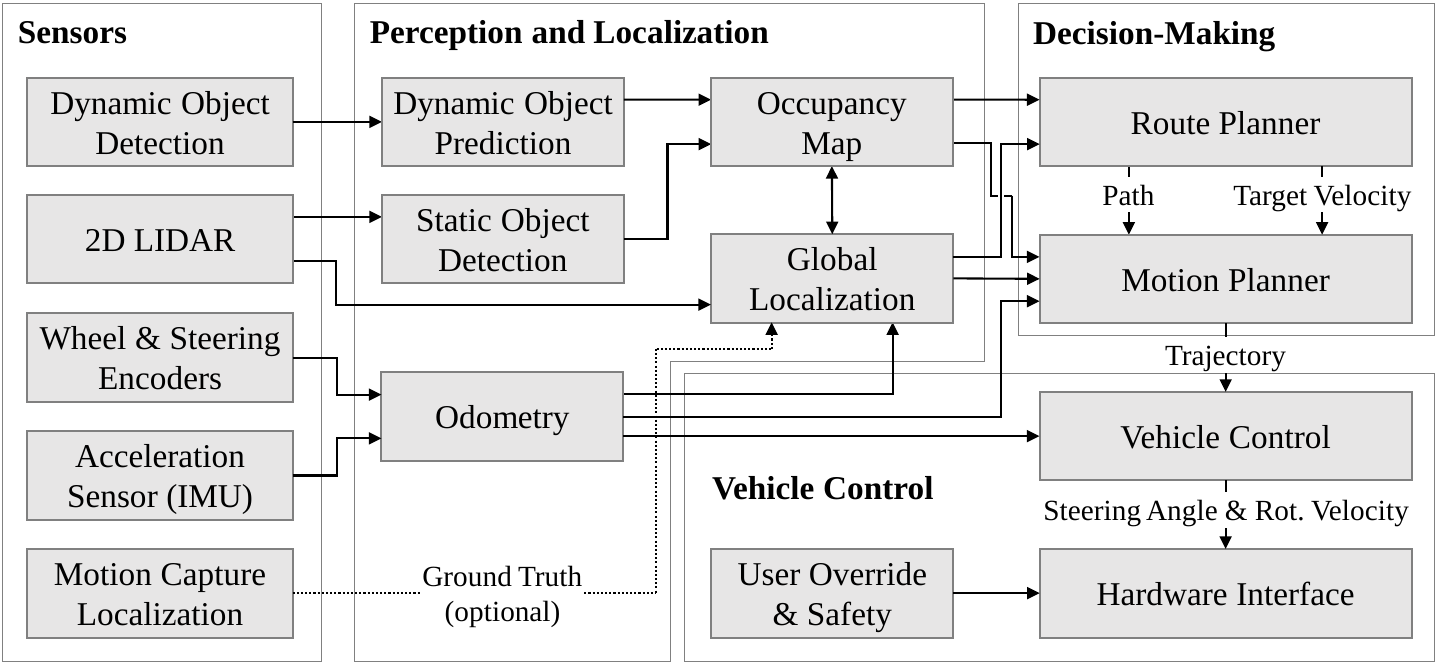}
      \caption{Functional Architecture of \zelos.}
      \label{fig:automation_architecture}
\end{figure}

The implemented automation features a centralized automation architecture and serves as a baseline for validations. The automation is composed of multiple automation modules, each automation module functions as a self-contained software unit with defined interfaces to other automation modules. The communication between the automation modules is based on the \gls{ROS} \cite{quigley.2009}, which allows easy exchangeability of individual automation modules. The automation modules are categorized into sensors, perception and localization, decision-making, and vehicle control. \autoref{fig:automation_architecture} depicts the automation architecture and the interaction between the automation modules.

The modular design of the automation architecture is beneficial for the use-case of a research platform, as validating new methods only requires modifying or replacing the respective automation module. This particularly reduces the effort required for onboarding and implementation.
\subsection{Safety Concept}
\zelos\ features an integrated systemwide safety concept. Each drive module publishes its status to a central safety watchdog on the main module. If any of the drive modules encounters an actuator or communication error, the watchdog instantly relays a safe-stop event to all drive modules, forcing \zelos\ into a safe state. In addition, the watchdog monitors time-stamped communication messages and forces a safe stop of all drive modules if it does not receive a safety-relevant message within a defined timeout. Conversely, if a drive module no longer receives commands from the main module within the timeout, it triggers a safe stop on its own. Furthermore, a safe stop can be triggered by any automation module and via the user interface.

In addition, \zelos\ features a safety system for removing drive torque (\gls{STO}). This functionality provides an independent second layer of fault response. It can be triggered via a certified wireless emergency stop system and a hardware emergency stop switch on \zelos.

\section{Modeling of the Research Platform}\label{sec:vehicle_model}
In this section, we present the model of \zelos, which is based on \cite{moseberg.2016}. We assume a flat surface for driving. For this reason, we neglect the cross-coupling of the horizontal and vertical dynamics and only model the horizontal dynamics. Moreover, as the powerful actuators allow for setting a desired tire steering angle $ \delta\indRunI $ and angular velocity $ \omega\indRunI $ of a tire $ i \in \cBrack[\text{fl, fr, rl, rr}] $ almost instantly (see \autoref{sec:actuators}), we neglect the actuator dynamics. Thus, the input of \zelos\ is
\begin{equation}\label{eq:zelos_input}
    \vu_\mathrm{ZeloS} = \begin{bmatrix}
        \delta\indfl & \delta\indfr & \delta\indrl & \delta\indrr & \omega\indfl & \omega\indfr & \omega\indrl & \omega\indrr
    \end{bmatrix}\transpose.
\end{equation}
The meaning of the indices is as follows: $ \mathrm{fl} $: front left, $ \mathrm{fr} $: front right, $ \mathrm{rl} $: rear left, and $ \mathrm{rr} $: rear right.
\subsection{Tire Model}\label{sec:tire_model}
Each tire generates a longitudinal (index $ \sx $) and lateral (index $ \sy $) force $ \vF\indRunI = \begin{bmatrix} \sF\indRunIx & \sF\indRunIy \end{bmatrix}\transpose $ %,  $ i \in \cBrack[\text{fl, fr, rl, rr}] $
w.r.t. the vehicle body. This force depends on the translational velocity $ \vv\indRunI = \begin{bmatrix} \sv\indRunIx & \sv\indRunIy \end{bmatrix}\transpose $, the steering angle $ \delta\indRunI $, and the rotational velocity $ \omega\indRunI $ of the respective tire. The force $ \vF\indRunI $ can be controlled by the input $ \vu_\mathrm{ZeloS} $. We use the simplified \textit{Magic Formula Tire Model} presented in \cite[Sec. 2.4.1]{moseberg.2016} to determine $ \vF\indRunI $: for brevity, assume $ \vF\indRunI = \vf_\mathrm{tire}\rBrack[\omega\indRunI,\delta\indRunI,\vv\indRunI] $.
\subsection{Tire Chassis Coupling}
The coupling between the four tires and the \gls{COG} of \zelos\ is constant (see \autoref{sec:body_design}) and given by the matrix 
\begin{equation}\label{eq:mat_G}
	\vG = 
	{\footnotesize
	\begin{bmatrix}
		1		&	  0   &    1    &  0    &1       &0      &1     &0\\
		0		&	  1   &    0    &1      &0       &1      &0     &1\\ 
		-s\indfl	&	l\indfl   & s\indfr    & l\indfr   &-s\indrl   &-l\indrl   &s\indrr  &-l\indrr
	\end{bmatrix}
	}.
\end{equation}
The parameter $ s\indRunI $ indicates a lateral distance, and the parameter $ l\indRunI $ indicates a longitudinal distance between a tire contact point and the \gls{COG} with $ i \in \cBrack[\text{fl, fr, rl, rr}] $. The relation between the force $ \vF\indRunI $ generated by a tire and the force 
\begin{equation}
    \vF\indcog =  \begin{bmatrix} \sF\indx & \sF\indy & \sM\indz \end{bmatrix}\transpose
\end{equation}
that acts on \zelos' \gls{COG} ($ F\indx $ and $ F\indy $ refer to the longitudinal and lateral force, $ M\indz $ refers to the yaw torque) is given by
\begin{equation}\label{eq:f_cog}
 	\vF\indcog = \vG \vF\indxy,
\end{equation}
with the vector $ \vF\indxy = \begin{bmatrix} \vF\indfl\transpose & \vF\indfr\transpose & \vF\indrl\transpose & \vF\indrr\transpose \end{bmatrix}\transpose $. Equivalently, the relation between the translational velocity of the tire contact points $ \vv\indRunI $ and the velocity
\begin{equation}\label{eq:vv}
	\vv\indcog =
	\begin{bmatrix}
		\sv\indx & \sv\indy & \sdo
	\end{bmatrix}\transpose
\end{equation}
of \zelos' \gls{COG} ($ \sv\indx $ and $ \sv\indy $ refer to the longitudinal and lateral velocity, $ \sdo $ refers to the yaw rate) is given by 
\begin{equation}\label{eq:v_xy}
    \vv_\mathrm{xy} = \vG\transpose \vv\indcog,
\end{equation}
with the vector $ \vv\indxy = \begin{bmatrix} \vv\transpose\indfl & \vv\indfr\transpose & \vv\indrl\transpose & \vv\indrr\transpose \end{bmatrix}\transpose $.
\subsection{Horizontal Dynamics}
The frame, in which $ \vF\indcog $ and $ \vv\indcog $ are defined, is affixed to \zelos' \gls{COG} (so-called body frame, see \autoref{fig:frames}). Therefore, note that this frame is not an inertial system.

The force $ \vF\indcog $ induces an acceleration
\begin{equation}\label{eq:a2f}
    \va\indcog = 
    \begin{bmatrix}
		\sa\indx \\ \sa\indy \\ \sddo
	\end{bmatrix}
    =
    \begin{bmatrix}
        \tfrac{1}{m} & 0 & 0 \\
        0 & \tfrac{1}{m} & 0 \\
        0 & 0 & \tfrac{1}{J\indz}
    \end{bmatrix}
    \vF\indcog
\end{equation}
of \zelos' \gls{COG}, $ \sa\indx $ and $ \sa\indy $ refer to the longitudinal and lateral acceleration, $ \sddo $ refers to the yaw acceleration. The parameter $ m $ represents \zelos' mass, and $ J\indz $ represents the rotational inertia of \zelos\ w.r.t. the vertical axis. The acceleration $ \va\indcog $ induces a change of \zelos' velocity according to
\begin{equation}\label{eq:hor_dyn}
    \dot{\vv}\indcog = \dFull{}{\stime}
    \begin{bmatrix}
        \sv\indx \\
        \sv\indy \\
        \sdo
    \end{bmatrix}
    =
    \underbrace{
    \begin{bmatrix}
        \phantom{-} \sv\indy \sdo \\
        -           \sv\indx \sdo \\
        0
    \end{bmatrix}}_{\bm{A}_\mathrm{v}\rBrack[{\vv\indcog}]}
    +
    \underbrace{
    \begin{bmatrix}
        1 & 0 & 0 \\
        0 & 1 & 0 \\
        0 & 0 & 1
    \end{bmatrix}}_{\bm{B}_\mathrm{v}}
    \va\indcog.
\end{equation}
Finally, the velocity $ \vv\indcog $ causes a change in the position
\begin{equation}\label{eq:vx}
    \vx\indcog = \begin{bmatrix} \sx & \sy & \so \end{bmatrix}\transpose
\end{equation}
of \zelos' \gls{COG}. The position $ \vx\indcog $ is defined in a global frame and changes according to
\begin{equation}\label{eq:vx_comp}
    \vxdot\indcog = \dFull{}{\stime}
    \begin{bmatrix}
        \sx \\
        \sy \\
        \so
    \end{bmatrix}
    =
    \begin{bmatrix}
        \cos\rBrack[\so] & -\sin\rBrack[\so] & 0\\
        \sin\rBrack[\so] & \phantom{-}\cos\rBrack[\so] & 0\\
        0 & 0 & 1
    \end{bmatrix} \vv\indcog.
\end{equation}

\section{Sensor Setup and Localization}
In this section, we describe the sensor setup and the implemented localization methods. Each module can be replaced by a different module that fulfills the same task. In the following, a hat on a variable indicates a measured value, e.g., $ \hat{\vx}\indcog $ indicates the measured value of $ \vx\indcog $.
\subsection{Coordinate Frames}\label{sec:frames}
\begin{figure}[t!]
    \centering
    \includegraphics[
 width=0.7\columnwidth,
 height=6cm,
 keepaspectratio,
 ]{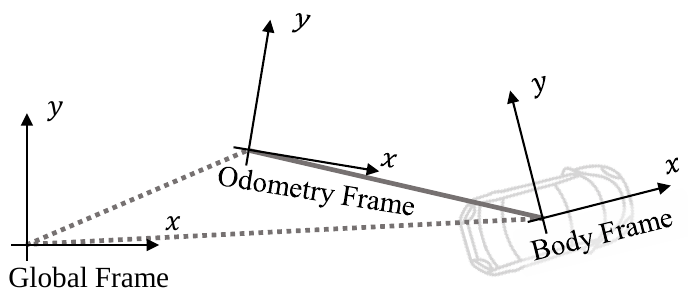}
    \caption{Coordinate frames used for the localization. Solid gray line: continuous transformation, dotted gray lines: discontinuous transformations.}
    \label{fig:frames}
\end{figure}

Three coordinate frames are used for the localization: a global frame, an odometry frame, and a body frame (see \autoref{fig:frames}). The global frame is static w.r.t. a map, and the body frame is affixed to \zelos' \gls{COG}. The localization aims to accurately determine the transformation between the body frame and the global frame at a high frequency. However, this transformation can only be determined accurately at a low frequency (e.g., using a 2D LIDAR-based method, see \autoref{sec:cartographer}). To resolve this, a high-frequency but less accurate localization of the body frame in an odometry frame is obtained (e.g., via encoder-based odometry, see \autoref{sec:odometry}). To ensure a high level of accuracy is still maintained, the accurate but low-frequency localization corrects the inaccuracy of the high-frequency but less accurate localization by adjusting the origin of the odometry frame in the global frame. This approach prevents discontinuities in the transformation between the body frame and the odometry frame, which is essential for vehicle control \cite{werling.2017}.% The discontinuous adjustments of the odometry frame are handled by the motion planner.
\subsection{Encoder-based Odometry}\label{sec:odometry}
The encoder-based odometry determines the velocity $ \hat{\vv}\indcog $ (see \eqref{eq:vv}) and the position $ \hat{\vx}\indcog $ (see \eqref{eq:vx}) of \zelos' \gls{COG} at a high frequency. For this, \zelos\ features relative encoders that measure the rotational velocity $ \hat{\omega}\indRunI $ and absolute encoders that measure the steering angle $ \hat{\delta}\indRunI $ of a tire. %$ i \in \cBrack[\text{fl, fr, rl, rr}] $. 
Because pure rolling of the tires (i.e., no tire slip) is assumed based on $ \hat{\omega}\indRunI $ and $ \hat{\delta}\indRunI $, inevitable inaccuracies in $ \hat{\vv}\indcog $ accumulate in $ \hat{\vx}\indcog $. To ensure a continuous localization while allowing for corrections of the inaccuracies, $ \hat{\vv}\indcog $ and $ \hat{\vx}\indcog $ are defined in the odometry frame (see \autoref{sec:frames} and \autoref{fig:frames}).

The translational velocities in $ \hat{\vv}\indRunI $ of a tire contact point are determined from the encoder measurements using
\begin{equation}
 \hat{\vv}\indRunI = \begin{bmatrix}
 \hat{\sv}\indRunIx \\ \hat{\sv}\indRunIy
    \end{bmatrix} = \begin{bmatrix}
 \hat{\omega}\indRunI r_\mathrm{dyn} \cos\rBrack[\hat{\delta}\indRunI] \\
 \hat{\omega}\indRunI r_\mathrm{dyn} \sin\rBrack[\hat{\delta}\indRunI]
    \end{bmatrix},
\end{equation}
$ r_\mathrm{dyn} $ is the tire radius. The velocity $ \hat{\vv}\indcog $ is determined based on \eqref{eq:v_xy}, using $ \hat{\vv}\indxy $, and thus, $ \hat{\vv}\indRunI $ and the Moore Penrose inverse matrix $ \overline{\vG}^+ $ of $ \vG\transpose $ (see \eqref{eq:mat_G}), 
\begin{equation}
 \hat{\vv}\indcog = \overline{\vG}^+ \hat{\vv}\indxy.
\end{equation}

The position $ \hat{\vx}\indcog $ is determined based on $ \hat{\vv}\indcog $ using \eqref{eq:vx_comp}.
\subsection{IMU Acceleration Sensor}
To mitigate the inaccuracies that result from neglecting the tire slip in the encoder-based odometry, an \gls{IMU} is used to determine $ \hat{\sa}\indx $, $ \hat{\sa}\indy $, and $ \hat{\sdo} $. This data is used in an Extended Kalman Filter for improving $ \hat{\vv}\indcog $ as determined by the encoder-based odometry.
\subsection{2D LIDAR-based Cartographer}\label{sec:cartographer}
The 2D LIDAR sensor serves two purposes. First, it is used to generate an occupancy map (see \autoref{fig:reference_path}) of the environment. Second, it is used to localize the body frame in the global frame and, thus, to correct inaccuracies that result from neglecting the tire slip in the encoder-based odometry.

The occupancy map is generated using the \gls{SLAM} algorithm implemented in the cartographer \gls{ROS} package \cite{hess.2016}. This algorithm iteratively matches a 2D LIDAR scan with a so-called submap constructed from previous scans. In doing so, the \gls{SLAM} algorithm determines the transformation between the body frame and the global frame, which is used to correct the inaccuracies of the encoder-based odometry.

In addition, the \gls{SLAM} algorithm complements submaps with data from 2D LIDAR scans. Complete submaps form a global occupancy map, which is stored for future localization of \zelos.
\subsection{Motion Capture System-based Ground Truth}\label{sec:vicon}
An external Vicon motion capture system can be used to support or validate the onboard localization methods. The motion capture system uses infrared cameras to track markers attached to \zelos\ for determining \zelos' position. This technology surpasses the accuracy of the onboard sensors. Thus, we consider the provided position to be ground truth.

\section{Decision-Making and Vehicle Control}
In this section, we present the implemented decision-making and control automation modules. Note that any module that fulfills the same task can replace the decision-making and control modules. Moreover, we provide some details on the implementation of the automation architecture.
\subsection{Scenario Generation}
We developed a custom scenario editor for generating reference paths. The tool visualizes recorded occupancy maps so that the user can select waypoints. In addition, a target velocity can be set for different parts of the reference path. The scenario editor automatically interpolates the waypoints and generates a reference path $ \mathfrak{X}\indref^\mathrm{p} $ which is defined by a set of reference entities $ \vx\indref^{\mathrm{p},j} = \begin{bmatrix}
    \sx\indref^\mathrm{p} & \sy\indref^\mathrm{p} & \so\indref^\mathrm{p} & \sv\indref^\mathrm{p}
\end{bmatrix} $, $ j = 1\dots N_\mathrm{s} $, with $ N_\mathrm{s} $ denoting the size of the reference path. The entries $ \sx\indref^\mathrm{p} $, $ \sy\indref^\mathrm{p} $, and $ \so\indref^\mathrm{p} $ are defined in the global frame and represent reference values for $ \sx\indcog $, $ \sy\indcog $, and $ \so\indcog $. Further, $ \sv\indref^\mathrm{p} $ represents the target velocity in the direction of $ \so\indref^\mathrm{p} $.
\subsection{Motion Planning}
The motion planning module generates a reference trajectory $ \mathfrak{X}\indref^\mathrm{t} $ that follows the reference path $ \mathfrak{X}\indref^\mathrm{p} $. The reference trajectory $ \mathfrak{X}\indref^\mathrm{t} $ is defined as a set of reference entities $ \vx\indref^{\mathrm{t},k} = \begin{bmatrix}
    \vx\indref & \vv\indref & \va\indref & \stime\indref
\end{bmatrix} $, with $ \vx\indref = \begin{bmatrix}
    \sx\indref & \sy\indref & \so\indref
\end{bmatrix}\transpose $, $ \vv\indref = \begin{bmatrix}
    \sv\indxref & \sv\indyref & \sdo\indref
\end{bmatrix}\transpose $, and $ \va\indref = \begin{bmatrix}
    \sa\indxref & \sa\indyref & \sddo\indref
\end{bmatrix}\transpose $ being the reference values for $ \vx\indcog $, $ \vv\indcog $, $ \va\indcog $. The time $ \stime\indref $ represents the time point for which $ \vx\indref $, $ \vv\indref $, and $ \va\indref $ are valid. 

The position values $ \vx\indref $ of the reference trajectory are defined in the odometry frame. Recalling that the reference positions of the reference path are defined in the global frame, the localization corrections provided by the cartographer (see \autoref{sec:cartographer}) are handled by the motion planning.

The implemented motion planning algorithm is a model predictive planning algorithm that cyclically solves a convex optimization problem. For motion planning, we consider the horizontal dynamics of \zelos' \gls{COG} with $ \sv\indy = 0 $ and $ \sa\indy = 0 $. The state vector of the planning model is $ \vx\indP \in \admissX\indP \subseteq \reals[5] $ with $ \vx\indP = \begin{bmatrix}
    \sx & \sy & \so & \sv\indx & \sdo
\end{bmatrix}\transpose $ and the input vector is $ \vu\indP \in \admissU\indP \subseteq \reals[2] $ with $ \vu\indP = \begin{bmatrix}
    \sa\indx & \sddo
\end{bmatrix}\transpose $. Both $ \admissX\indP $ and $ \admissU\indP $ are assumed to be convex sets. The dynamics of the planning model are given by $ \vxdot\indP = \vf\indP\rBrack[\vx\indP,\vu\indP] $ with
\begin{equation}\label{motplan:nonlinear_model}
    \vf\indP\rBrack[\vx\indP,\vu\indP] = \begin{bmatrix}
        \sv\indx \cos\rBrack[\so]\\
        \sv\indx \sin\rBrack[\so]\\
        \sdo\\
        \sa\indx\\
        \sddo
    \end{bmatrix}.
\end{equation}
To ensure that the optimization problem related to motion planning can be solved efficiently, \eqref{motplan:nonlinear_model} is linearized along the previously planned solution trajectory $ \mathfrak{X}\indref^\mathrm{t,prev} $. % $ \vx_\mathrm{p,sol}^\mathrm{prev} $. This solution trajectory results from applying the optimal control trajectory $ \vu_\mathrm{p,sol}^\mathrm{prev} $ that has been determined in the previous planning cycle to the nonlinear vehicle model \eqref{motplan:nonlinear_model}. 
The resulting linear system is discretized w.r.t. time, which results in
\begin{equation}
    \vx\indP^{k+1}=\bm{A}\indP^k \vx\indP^k + \bm{B}\indP^k \vu\indP^k,
\end{equation}
$ k = 1 \dots N_\mathrm{p} $, in which $ N_\mathrm{p} $ is the size of the prediction horizon. To ensure the convexity of the optimization problem, the admissible solution space $ \admissX\indPA \subseteq \admissX\indP $ must remain convex when considering obstacles that must be avoided in the $ \sx $-$ \sy $-plane. This is ensured using the algorithm presented in \cite{zhong.2020} for finding large convex polytopes directly on the point cloud determined by the 2D LIDAR sensor. The resulting convex, collision-free polytope is reduced by the radius of a circle that encloses \zelos' vehicle body to avoid collisions. This yields the convex admissible solution space $ \admissX\indPA $.

The motion planning and control modules of \zelos\ interact in a bi-level architecture as presented in \cite{werling.2011}. This bi-level architecture is known for well-handling model errors, permanent disturbances, and impulse disturbances \cite{werling.2011}. If the absolute value of the tracking error $ \vx\indeps^\mathrm{a} = \abs[{\vx_\mathrm{p,sol}^{\mathrm{prev},k=2} - \hat{\vx}\indP}] $ is smaller than a threshold vector $ \vx\indeps^\mathrm{th} = \begin{bmatrix}
 \sx\indeps^\mathrm{th} & \sy\indeps^\mathrm{th} & \so\indeps^\mathrm{th} & \sv\indeps^\mathrm{th} & \sdo\indeps^\mathrm{th}
\end{bmatrix}\transpose $, planning is performed open-loop. Only if $ \vx\indeps^\mathrm{a} > \vx\indeps^\mathrm{th} $, the initial state vector for the next planning cycle is reinitialized at $ \hat{\vx}\indP $.

The convex optimization problem for motion planning is formulated by means of two quadratic cost terms. The cost term $ J\indref^k\rBrack[\vx\indP^k,\vu\indP^k] $ ensures that the resulting reference trajectory follows the reference path at the target velocity, and $ J_\mathrm{obs}^k\rBrack[\vx^k\indP] $ supports the motion planner to avoid obstacles. The optimization problem is given by:
\begin{subequations}
    \begin{equation}
        \min_{\vu\indP^1\dots\vu\indP^{N_\mathrm{p}-1}} \sum_{k=1}^{N_\mathrm{p}}J\indref^k\rBrack[\vx\indP^k,\vu\indP^k] + J_\mathrm{obs}^k\rBrack[\vx^k\indP],
    \end{equation}
    \begin{center}
        subject to
    \end{center}
    \begin{equation}
        %\vxdot\indP = \vf\indP\rBrack[\vx\indP,\vu\indP]
        \vx\indP^{k+1}=\bm{A}\indP^k \vx\indP^k + \bm{B}\indP^k \vu\indP^k,
    \end{equation}
    \begin{equation}
        \vx\indP^k \in \admissX\indPA,
    \end{equation}
    \begin{equation}
        \vu\indP^k \in \admissU\indP,
    \end{equation}
    \begin{equation}
        \vx\indP^{k=1} = \left\{\begin{matrix*}[l]
            \vx_\mathrm{p,sol}^\mathrm{prev,k=2} &\text{ if } \vx\indeps^\mathrm{a} \leq \vx\indeps^\mathrm{t},\\
            \hat{\vx}\indcog  &\text{ if } \vx\indeps^\mathrm{a} > \vx\indeps^\mathrm{t}.
        \end{matrix*}\right.
    \end{equation}
\end{subequations}

The result of this optimization problem yields $ \mathfrak{X}\indref^\mathrm{t} $. The planning algorithm is executed at a rate of $ f_\mathrm{p} = \SI{5}{\hertz} $. The prediction time of the planning algorithm is $ \stime_\mathrm{p,pred} = \SI{4}{\second} $, the sample time of $ \mathfrak{X}\indref^\mathrm{t} $ is $ \Delta\stime_\mathrm{p} = \SI{200}{\milli\second} $, and thus, the resulting reference trajectory contains $ N_\mathrm{p} = 20 $ reference entities $ \vx\indref^{\mathrm{t},k} $. Before sending this reference trajectory to the tracking controller, it is interpolated to $ \Delta\stime_\mathrm{c} = \SI{10}{\milli\second} $.
\subsection{Tracking Control}
\begin{figure}[t!]
    \centering
    \includegraphics[
        width=\columnwidth,
        height=6cm,
        keepaspectratio,
      ]{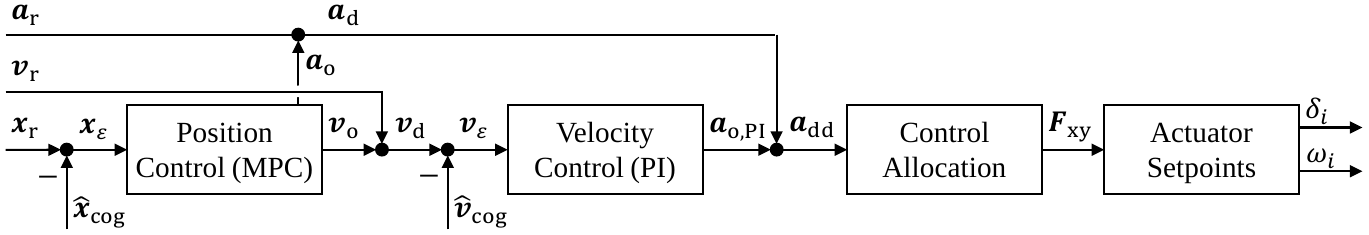}
      \caption{Architecture of \zelos' vehicle dynamics control.}
      \label{fig:control}
\end{figure}

The tracking control of \zelos\ is designed to explicitly consider input and state constraints. There are two control cascades: a position feedback control and a velocity feedback control. Additionally, the reference velocity $ \vv\indref $ and the reference acceleration $ \va\indref $ are used in feedforward controllers. The control architecture is depicted in \autoref{fig:control}.

In the following, the index $ \mathrm{d} $ refers to a \textit{desired} value that is valid for the \gls{COG}, which is determined by the position control and communicated to velocity control. A desired value is the sum of a reference value (index $ \mathrm{r} $) and a so-called offset value (index $ \mathrm{o} $) that is determined by the position feedback controller, that is $ \vv\inddes = \vv\indref + \vv\indoff $ and $ \va\inddes = \va\indref + \va\indoff $.
\subsubsection{Position Control}
We use \gls{MPC} in a feedback control architecture for tracking the reference position $ \vx\indref $. An additional feedforward controller uses the reference velocity $ \vv\indref $ and reference acceleration $ \va\indref $ to control the setpoint profile of the reference trajectory. The position tracking error $ \vx\indeps = \begin{bmatrix}
    \serr\indx & \serr\indy & \serr\indo
\end{bmatrix}\transpose $ is formulated in the body frame and is composed of the longitudinal error $ \sx\indeps $, the lateral error $ \sy\indeps $, and the orientation error $ \so\indeps $, 
\begin{equation}\label{eq:pos_error}
    \vx\indeps = {\footnotesize\begin{bmatrix}
        \phantom{-}\cos\rBrack[\hat{\so}\indcog] & \sin\rBrack[\hat{\so}\indcog] & 0\\
        -\sin\rBrack[\hat{\so}\indcog] & \cos\rBrack[\hat{\so}\indcog] & 0\\
        0 & 0 & 1
    \end{bmatrix}
    \rBrack[{
        \begin{bmatrix}
            \sx\indref \\ \sy\indref \\ \so\indref
        \end{bmatrix}
        -
        \begin{bmatrix}
            \hat{\sx}\indcog \\ \hat{\sy}\indcog \\ \hat{\so}\indcog
        \end{bmatrix}
    }]}.
\end{equation}
Differentiating \eqref{eq:pos_error} yields the error dynamics $ \vxdot\indeps = \dFullInline{\vx\indeps}{t} $,
\begin{equation}\label{eq:err_dyn}
    \vxdot\indeps = 
    {\footnotesize\begin{bmatrix*}[l]
        \phantom{-} \sdo\inddes \cdot \serr\indy - \sv\indxdes + \sv\indxref \cos\rBrack[{\serr\indo}] - \sv\indyref \sin\rBrack[{\serr\indo}] \\
        -           \sdo\inddes \cdot \serr\indx - \sv\indydes - \sv\indxref \sin\rBrack[{\serr\indo}] + \sv\indyref \cos\rBrack[{\serr\indo}]\\
        \phantom{-} \sdo\inddes - \sdo\indref
    \end{bmatrix*}}.
\end{equation}
The input of system \eqref{eq:err_dyn} is given by the desired velocity $ \vv\inddes $. For considering the feedforward controllers in the \gls{MPC}, we substitute $ \vv\inddes = \vv\indref + \vv\indoff $. With this, $ \vv\indoff = \begin{bmatrix}
    \sv_\mathrm{o,\sx} & \sv_\mathrm{o,\sy} & \sdo_\mathrm{o}
\end{bmatrix}\transpose $ is the input of system \eqref{eq:err_dyn}.

To constrain the offset acceleration $ \va\indoff $, the \gls{MPC} considers the state vector $ \vx_\mathrm{c} = \begin{bmatrix}
    \vx\indeps\transpose & \vv\indoff\transpose
\end{bmatrix}\transpose $. The dynamics of $ \dot{\vv}\indoff = \dFullInline{\vv\indoff}{t} $ are given by
\begin{equation}\label{eq:dyn_vo}
    \dot{\vv}\indoff = {\footnotesize\begin{bmatrix*}[l]
        \sa\indxoff + \rBrack[{\sdo\indref + \sdo\indoff}] \cdot \rBrack[{\sv\indyref + \sv\indyoff}] - \rBrack[{\sdo\indref \cdot \sv\indyref}]\\
        \sa\indyoff - \rBrack[{\sdo\indref + \sdo\indoff}] \cdot \rBrack[{\sv\indxref + \sv\indxoff}] + \rBrack[{\sdo\indref \cdot \sv\indxref}]\\
        \sddo\indoff
    \end{bmatrix*}},
\end{equation}
with the input $ \va\indoff = \begin{bmatrix}
    \sa\indxoff & \sa\indyoff & \sddo_\mathrm{o}
\end{bmatrix}\transpose = \va\inddes - \va\indref $.

Thus, the \gls{MPC} uses the input $ \va\indoff $ to control the system
\begin{equation}\label{eq:mpc_sys}
    \vxdot_\mathrm{c} = \begin{bmatrix}
        \vxdot\indeps \\ \dot{\vv}\indoff
    \end{bmatrix}
    =
    \begin{bmatrix}
        \textit{see \eqref{eq:err_dyn}}\\
        \textit{see \eqref{eq:dyn_vo}}\\
    \end{bmatrix}.
\end{equation}

For efficiently solving the \gls{MPC} optimization problem, \eqref{eq:mpc_sys} is linearized around the optimal solution of the previous \gls{MPC} cycle and discretized w.r.t. time. This results in% $ \vxdot_\mathrm{c} = \vf_\mathrm{lin}\rBrack[\vx_\mathrm{c},\va\indoff] $.
\begin{equation}\label{eq:lin_planning_sys}
    \vx_\mathrm{c}^{l+1}=\bm{A}_\mathrm{c}^l \vx_\mathrm{c}^l + \bm{B}_\mathrm{c}^l \vu_\mathrm{c}^l,
\end{equation}
$ l = 1\dots N_\mathrm{c} $, $ N_\mathrm{c} $ denotes the size of the prediction horizon. The \gls{MPC} constrains the offset velocity $ \vv\indoff $ to a polytopic set $ \mathcal{V}_\mathrm{o} \subseteq \reals[3] $ and the offset acceleration $ \va\indoff $ to a polytopic set $ \mathcal{A}_\mathrm{o} \subseteq \reals[3] $. The sets $ \mathcal{V}_\mathrm{o} $ and $ \mathcal{A}_\mathrm{o} $ can be defined depending on the given reference values $ \vv\indref $ and $ \va\indref $. Thus, the \gls{MPC} is capable of constraining $ \vv\inddes $ and $ \va\inddes $.

Finally, the \gls{MPC} optimization problem is defined using the matrices $ \bm{Q} \in \reals[6\times6] $, $ \bm{S} \in \reals[6\times6] $, and $ \bm{R} \in \reals[3\times3] $:
\begin{subequations}
    \begin{equation}
        \min_{\va\indoff^1\dots\va\indoff^{N_\mathrm{c}-1}} {\vx_\mathrm{c}^{N_\mathrm{c}}}\transpose \bm{S} \vx_\mathrm{c}^{N_\mathrm{c}} + \sum_{l=1}^{N_\mathrm{c}-1} {\vx_\mathrm{c}^l}\transpose \bm{Q} \vx_\mathrm{c}^l + {\va\indoff^l}\transpose \bm{R} \va\indoff^l,
    \end{equation}
    \begin{center}
        subject to
    \end{center}
    \begin{equation}
        %\vx_\mathrm{c}^{l+1} = \vf_\mathrm{lin}\rBracl[\vx_\mathrm{c}^l,\va\indoff^l]
        \vx_\mathrm{c}^{l+1}=\bm{A}_\mathrm{c}^l \vx_\mathrm{c}^l + \bm{B}_\mathrm{c}^l \vu_\mathrm{c}^l,
    \end{equation}
    \begin{equation}
        \vv\indoff^l \in \mathcal{V}_\mathrm{o},
    \end{equation}
    \begin{equation}
        \va\indoff^l \in \mathcal{A}_\mathrm{o},
    \end{equation}
    \begin{equation}
        \vx_\mathrm{c}^{l=1} = \begin{bmatrix}
            \hat{\vx}\indcog\transpose & {\vv_\mathrm{o,sol}^{\mathrm{prev},l=2}}\transpose
        \end{bmatrix}\transpose,
    \end{equation}
\end{subequations}
in which $ \vv_\mathrm{o,sol}^{\mathrm{prev},l=2} $ refers to the second value of the optimal solution that has been determined in the previous \gls{MPC} cycle.

The \gls{MPC} is executed at a cycle rate of $ f_\mathrm{c} = \SI{100}{\hertz} $, the prediction time is $ t_\mathrm{c,pred} = \SI{1}{\second} $, the sample time is $ \Delta\stime_\mathrm{c} = \SI{10}{\milli\second} $, and thus, the prediction horizon contains $ N_\mathrm{c} = 100 $ samples. If one optimization does not finish within the \gls{MPC} cycle time, the shifted optimal solution that has been determined in the previous \gls{MPC} cycle is used.
\subsubsection{Velocity Control}
For controlling the horizontal dynamics \eqref{eq:hor_dyn}, the velocity control designed in \cite{schwartz.2020} is implemented. The output $ \bm{y} = \vv\indcog $ of \eqref{eq:hor_dyn} is differentially flat. Therefore, the feedforward part $ \vu_\mathrm{v,ff} $ of the velocity control is (see \cite{schwartz.2020})
\begin{equation}
    \vu_\mathrm{v,ff} = \bm{B}_\mathrm{v}^{-1}\rBrack[{\dot{\vv}\inddes - \bm{A}_\mathrm{v}}\rBrack[{\vv\inddes}]] = \va\inddes = \va\indref + \va\indoff.
\end{equation}

A PI controller determines the feedback velocity control $ \vu_\mathrm{v,fb} $. Using the proportional $ \vR_\mathrm{P} = \mathrm{diag}\rBrack[ R_\mathrm{P,\sx}, R_\mathrm{P,\sy},
R_\mathrm{P,\so}] $ and integral $ \vR_\mathrm{I} = \mathrm{diag}\rBrack[ R_\mathrm{I,\sx}, R_\mathrm{I,\sy},R_\mathrm{I,\so}] $ gains yields
\begin{equation}
    \vu_\mathrm{v,fb} = \va_\mathrm{o,PI} = \vR_\mathrm{P} \rBrack[{\vv\inddes - \hat{\vv}\indcog}] + \vR_\mathrm{I} \int \rBrack[{\vv\inddes - \hat{\vv}\indcog}] \diff \tau.
\end{equation}

The desired acceleration $ \va_\mathrm{dd} $ for the \gls{COG} results from
\begin{equation}\label{eq:vel_ctrl}
    \va_\mathrm{dd} = \vu_\mathrm{v,ff} + \vu_\mathrm{v,fb} = \va\inddes + \va_\mathrm{o,PI}.
\end{equation}
\subsubsection{Control Allocation}
Equation \eqref{eq:a2f} yields the desired force $ \vF\inddes $ for the \gls{COG} based on $ \va_\mathrm{dd} $. The control allocation allocates the desired force $ \vF\inddes $ to the four tires. The force desired to be generated by a tire is $ \vF_\mathrm{xy,d} $. We use the analytical control allocation designed in \cite[Sec. 5.1]{moseberg.2016}:%, which is based on the inverse tire chassis coupling
\begin{equation}
    \vF_\mathrm{xy,d} = \vG^+ \vF\inddes + \vG^\perp \Delta\vF\indxy.
\end{equation}
The matrix $ \vG^+ $ is the Moore Penrose inverse of $ \vG $ (see \eqref{eq:mat_G}), $ \vG^\perp $ is such that $ \vG\vG^\perp = \bm{0} $, and $ \Delta\vF\indxy $ contains parameters for allocating the forces between the tires.
\subsubsection{Actuator Setpoint Generation}
For a tire to generate the force $ \vF_\mathrm{xy,d} $, the required tire steering angle $ \delta $ and rotational velocity $ \omega $ must be determined. For this, the procedure presented in \cite[Sec. 4.2.1]{moseberg.2016} is used, which relies on the analytical inversion of the simplified \textit{Magic Formula Tire Model} (see \autoref{sec:tire_model}). This yields $ \vu_\mathrm{ZeloS} $ (cf. \eqref{eq:zelos_input}), which is sent to the drive control units (see \autoref{sec:hw_interface}).
\subsection{Implementation}
All automation modules are implemented in C++, except for the drive control unit, which is implemented in C. To ensure reliable communication between the automation modules, each module features a \gls{ROS} wrapper. All modules are timed cyclic asynchronously, most of them at $ f = \SI{100}{\hertz} $.

%The motion planning and control modules feature optimization-based methods. 
The motion planning optimization problem is solved using IPOPT \cite{wachter.2006}, and the position \gls{MPC} optimization problem is solved using HPIPM \cite{frison.2020}. CasADi \cite{andersson.2019} performs the algorithmic differentiations and time discretizations are performed using the Runge-Kutta 4th-order integration scheme.

Note that a significant part of the implementation effort is required for the environment setup, e.g., consistent logging, synchronized time across all computational units, etc.

\section{Digital Twin for Simulation}
For validating new methods, a performant implementation of the respective method is necessary. For fast and risk-free testing of implementations and for testing the interaction of new methods with other automation modules, we use a digital twin of \zelos. The digital twin allows for software in the loop testing by providing the same interfaces as \zelos, realistic sensor behavior, and accurate dynamic behavior. The simulation is based on Gazebo \cite{koenig.2004}. However, the dynamic behavior of \zelos\ is simulated separately, as Gazebo does not easily simulate customized dynamics such as tire slip.

The model presented in \autoref{sec:vehicle_model} is complemented with adjustable latencies and actuator dynamics to simulate the dynamic behavior of \zelos. The simulated values of $ \vx\indcog $, $ \vv\indcog $, and $ \va\indcog $ are sent to Gazebo cyclically. With the updated position of \zelos\ in Gazebo, plugins are used for simulating the 2D LIDAR sensor, the \gls{IMU} sensor, and the motion capture system. Moreover, noise is added to the states of the actuators to simulate realistic encoder behavior.
\section{Experiments}
\begin{figure}[t!]
    \centering
    \input{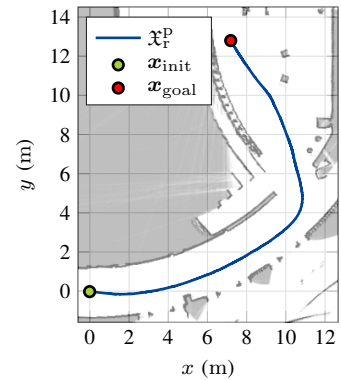}
    \caption{Occupancy map and reference path of the presented scenario, drivable areas are depicted in white.}% $ \vx\indref^\mathrm{path} $ that starts at the initial position $ \vx_\mathrm{init} $ and ends at the goal position $ \vx_\mathrm{goal} $.}
    \label{fig:reference_path}
\end{figure}

In this section, we present experimental data recorded in a scenario in which \zelos\ delivered the chain of office onto the stage at the inauguration ceremony of the new president of the Karlsruhe Institute of Technology. \autoref{fig:reference_path} depicts the reference path of the presented scenario and the occupancy grid that has been generated with the cartographer.

\begin{figure}[t!]
    \centering
    \input{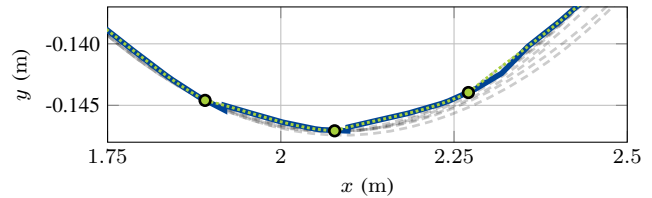}
    \caption{The blue line depicts the reference path in the odometry frame. The green dots depict the initial positions of the reference trajectories, the dotted green lines depict the initial segments of the reference trajectories, and the dashed gray lines depict the tails of the reference trajectories. The concatenated initial segments constitute the resulting reference trajectory.}
    \label{fig:reference}
\end{figure}
\autoref{fig:reference} depicts a segment of the reference path and trajectory in the odometry frame. The reference path is continuous in the global frame. However, in the odometry frame, discontinuities can be seen at $ \sx \approx \SI{1.9}{\meter} $ and $ \sx \approx \SI{2.1}{\meter} $. The discontinuities indicate the corrections of the odometry frame based on the localization of the cartographer (see \autoref{sec:cartographer}). Nevertheless, the resulting reference trajectory (dotted green line) is continuous, which indicates the benefit of defining the reference trajectory in the odometry frame.

\begin{figure}[t!]
    \centering
    \input{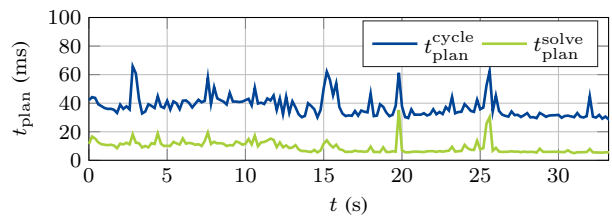}
    \caption{Cycle times and solving times of the motion planner.}
    \label{fig:t_cycle_planner}
\end{figure}
\begin{figure}[t!]
    \centering
    \input{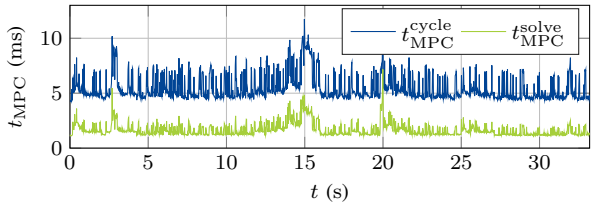}
    \caption{Cycle times and solving times of the \gls{MPC} position control.}
    \label{fig:t_cycle_mpc}
\end{figure}
\begin{figure}[t!]
    \centering
    \input{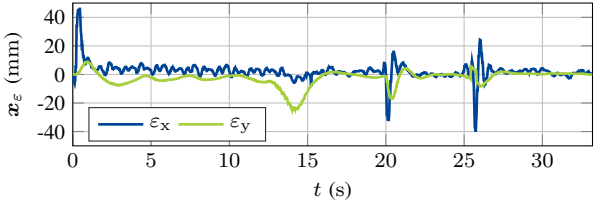}
    \caption{Position tracking error $ \vx\indeps $ in $ \sx $ and $ \sy $ direction (body frame).}
    \label{fig:tracking_error}
\end{figure}

\autoref{fig:t_cycle_planner} and \autoref{fig:t_cycle_mpc} depict the cycle times of the motion planning and position control. The times are measured on an Intel\textregistered\ Core\texttrademark\ i7-1185G7 CPU with $ \SI{4.8}{\giga\hertz} $ while all automation modules were running. The average cycle times in the presented scenario are $ \stime_\mathrm{plan}^\mathrm{cycle,avg} = \SI{37.54}{\milli\second} $ and $ \stime_\mathrm{MPC}^\mathrm{cycle,avg} = \SI{5.43}{\milli\second} $. All motion planning cycles and more than $ \SI{99.7}{\percent} $ of the \gls{MPC} cycles met the target cycle time in the presented scenario. As for the \gls{MPC} cycles that did not meet the target cycle time of $ \stime_\mathrm{MPC}^\mathrm{cycle,tar} < \SI{10}{\milli\second} $, the solution of the previous \gls{MPC} cycle was used. \autoref{fig:tracking_error} depicts the position tracking error, which is at most $ < \SI{20}{\milli\meter} $.

\begin{figure}[t!]
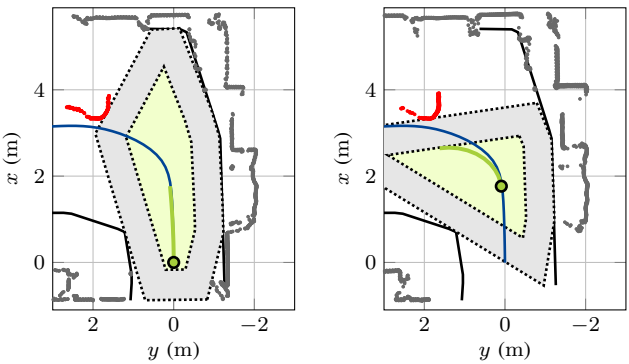

    \centering
    \begin{subfigure}[t]{0.49\columnwidth}
        \captionsetup{margin=2pt}
        \centering
       \input{02_fig/ltv_szenario_idx_3.tikz}
       \caption{Setting at $ \stime = \SI{0}{\second} $: the reference trajectory follows the reference path. An obstacle can be recognized close to the reference path, depicted in red.}
       \label{fig:planner1}
    \end{subfigure}
    \hfill
    \begin{subfigure}[t]{0.49\columnwidth}
        \captionsetup{margin=2pt}
        \centering
       \input{02_fig/ltv_szenario_idx_13_rec.tikz}
       \caption{Setting at $ \stime = \SI{2}{\second} $: to avoid a collision with the obstacle (depicted in red), the reference trajectory diverges from the reference path and stays inside $ \admissX\indPA $.}
       \label{fig:planner2}
    \end{subfigure}
    \caption{\zelos' position is at the green mark, the blue line depicts the reference path, and the green line depicts the reference trajectory. The gray dots depict noncritical points of the 2D LIDAR scan and the red dots depict critical points (obstacle), the solid black lines depict user-defined boundaries, and the light gray area depicts the convexified drivable area. The green area depicts the admissible solution space $ \admissX\indPA $.}
    \label{fig:planner}
\end{figure}
\autoref{fig:planner} shows data recorded in a different scenario with $ \stime_\mathrm{p,pred} = \SI{2}{\second} $. The figure depicts a segment of the environment as perceived by \zelos, and thus, the motion planning module. The figure illustrates the construction of the admissible solution space $ \admissX\indPA $ and demonstrates that the motion planning avoids an obstacle that is in \zelos' way.

\section{Conclusion}
In this paper, we present \zelos, a modular research platform that addresses both hardware and software requirements for validating automated driving methods in an early stage of research. The focus of this paper is on the implemented automation, particularly on motion planning and control. The modular hardware design of \zelos\ enables flexible reconfiguration of the wheel arrangement and switching between front-wheel, rear-wheel, and all-wheel steering. Moreover, the actuators provide performance reserves suitable for demanding validation tasks. The modular automation of \zelos\ is based on \gls{ROS} and allows for modifying or replacing individual automation modules with minimal effort. \zelos\ features a low-level safety concept that ensures reliable fault response by monitoring critical operations across all hardware and automation modules. The presented experimental results demonstrate that \zelos\ navigates reliably from an initial position to a target position and autonomously avoids obstacles. This highlights the suitability of \zelos\ for testing and validating a wide range of automated driving methods.

% References
\printbibliography

\end{document}